%% file: main.tex
\documentclass{article}

\PassOptionsToPackage{numbers}{natbib}

\usepackage[final]{neurips_2024}

\usepackage{float}

\usepackage[utf8]{inputenc} 
\usepackage[T1]{fontenc}    
\usepackage{hyperref}       
\usepackage{url}            
\usepackage{booktabs}       
\usepackage{longtable}      
\usepackage{amsfonts}       
\usepackage{nicefrac}       
\usepackage{microtype}      
\usepackage{xcolor}         
\usepackage{graphicx}       
\usepackage{multirow}
\usepackage{xcolor,colortbl}
\usepackage{adjustbox}
\usepackage{siunitx}
\usepackage[font=small]{caption}
\usepackage{placeins}
\usepackage{makecell}

\usepackage{seqsplit}       

\newcommand{\ttb}[1]{{\def\_{\textunderscore\allowbreak}\texttt{#1}}} 

\title{OpenTME: An Open Dataset of AI-powered H\&E Tumor Microenvironment Profiles from TCGA}
\author{
Maaike Galama$^{*\;1}$, 
Nina Kozar-Gillan$^{*\;1}$, 
Christina Embacher$^{*\;1}$, 
Todd Dembo$^{*\;1}$, \AND
Cornelius Böhm$^{1}$, 
Evelyn Ramberger$^{1}$, 
Julika Ribbat-Idel$^{1}$, 
Rosemarie Krupar$^{1}$, 
Verena Aumiller$^{1}$, \AND
Miriam Hägele$^{1}$, 
Kai Standvoss$^{1}$, 
Gerrit Erdmann$^{1}$, 
Blanca Pablos$^{1}$, 
Ari Angelo$^{1}$, \AND
Simon Schallenberg$^{2}$, 
Andrew Norgan$^{3}$, 
Viktor Matyas$^{1}$, 
Klaus-Robert Müller$^{4\;5\;6\;7}$, \AND
Maximilian Alber$^{\dag\;1\;2}$, 
Lukas Ruff$^{\dag\;1}$, 
Frederick Klauschen$^{\dag\;2\;5\;8\;9\;10}$ \\ \\
$^{1}$ Aignostics, Germany \AND 
$^{2}$ Institute of Pathology, Charité – Universitätsmedizin Berlin, Germany \AND 
$^{3}$ Department of Laboratory Medicine and Pathology, Mayo Clinic, Rochester, MN, US \AND 
$^{4}$ Machine Learning Group, Technische Universität Berlin, Germany \AND 
$^{5}$ BIFOLD – Berlin Institute for the Foundations of Learning and Data, Germany \AND 
$^{6}$ Department of Artificial Intelligence, Korea University, Republic of Korea \AND 
$^{7}$ Max-Planck Institute for Informatics, Germany \AND 
$^{8}$ German Cancer Research Center (DKFZ) \& German Cancer Consortium (DKTK), \\ Berlin \& Munich Partner Sites, Germany \AND 
$^{9}$ Institute of Pathology, Ludwig-Maximilians-Universität München, Germany \AND 
$^{10}$ Bavarian Cancer Research Center (BZKF), Germany \AND
${*}$ Equal contribution \AND
$\dag$ Corresponding author}

\begin{document}

\maketitle

\begin{abstract}
The tumor microenvironment (TME) plays a central role in cancer progression, treatment response, and patient outcomes, yet large-scale, consistent, and quantitative TME characterization from routine hematoxylin and eosin (H\&E)-stained histopathology remains scarce. 
We introduce OpenTME, an open-access dataset of pre-computed TME profiles derived from 5,268 H\&E-stained whole-slide images across ten cancer types (bladder, breast, colorectal, head and neck, liver, lung, ovarian, pancreatic, prostate, and stomach cancer) from The Cancer Genome Atlas (TCGA).
All outputs were generated using HETA (H\&E Tissue Analysis), an AI-powered application built on the Atlas family of pathology foundation models, which performs tissue quality control, tissue segmentation, cell detection and classification, tumor zone and tertiary lymphoid structure (TLS) detection, and spatial neighborhood analysis, yielding over 20,000 quantitative readouts per slide at cell-level resolution, including curated features such as tumor-infiltrating lymphocytes, tumor mitotic index, or mature TLS counts.
OpenTME is available for non-commercial academic research on Hugging Face. 
We will continue to expand OpenTME over time and anticipate it will serve as a resource for biomarker discovery, spatial biology research, and the development of computational methods for TME analysis.\footnote{OpenTME v2.0 is based on HETA v2.1. Latest releases are always available at: \url{https://huggingface.co/datasets/Aignostics/OpenTME}.}
\end{abstract}

\section{Introduction}
\label{sec:intro}

The tumor microenvironment (TME) --- comprising cancer cells, immune infiltrates, stromal components, vasculature, and their spatial organization --- is a central determinant of cancer biology, disease progression, and therapeutic response~\cite{saltz2018,wang2020}. 
Quantitative characterization of the TME from routine hematoxylin and eosin (H\&E)-stained histopathology slides has become an increasingly important avenue for biomarker discovery, patient stratification, and translational research. 
H\&E staining remains the most widely used preparation in diagnostic pathology, and the digitization of whole-slide images (WSIs) has opened the door to quantitative and computational analysis at scale. 

Recent advances in deep learning --- particularly the emergence of pathology foundation models~\cite{chen2024,zimmermann2024,alber2026} and advanced architectures for cell detection and tissue segmentation~\cite{graham2019,hoerst2024} --- have made it possible to extract detailed, cell-level TME profiles from H\&E images in a scalable and reproducible manner. 
However, applying such methods at cohort scale still requires significant computational infrastructure, domain 
expertise, and careful quality control. 
As a result, most researchers who could benefit from quantitative TME features --- for example, to correlate spatial immune patterns with genomic or clinical data --- often face a barrier to entry. 

To address this gap, we introduce \textbf{OpenTME}, an open-access dataset of pre-computed, quantitative TME profiles derived from 5,268 H\&E-stained diagnostic WSIs across ten cancer types (bladder, breast, colorectal, head and neck, liver, lung, ovarian, pancreatic, prostate, and stomach cancer) from The Cancer Genome Atlas (TCGA).
All outputs were generated using HETA (H\&E Tissue Analysis), an AI-powered application built on the Atlas family of pathology foundation
models~\cite{dippel2024,alber2025,alber2026}, which performs tissue quality control, tissue segmentation into nine tissue types, cell detection and classification into eleven cell classes, tumor zone and tertiary lymphoid structure (TLS) detection, and spatial neighborhood analysis --- yielding over 20,000 quantitative readouts per slide at single-cell resolution, including curated features that capture key biological characteristics of the TME.
The dataset is accompanied by TME Studio, a collection of interactive analysis notebooks to help researchers explore and build on the data. 

In addition to the OpenTME dataset, we introduce the \href{https://www.aignostics.com/academics#research-access-program}{\textbf{HETA Research Access Program}}, through which academic researchers can request to run HETA on their own data or to receive additional outputs such as complete cell coordinates, full model output polygon geometries, and segmentation maps beyond what is included in the open dataset.

By providing a ready-to-use, consistently generated TME characterization of a large, well-studied public cohort, we aim to lower the barrier to H\&E-based spatial biology research and to promote a broad range of downstream applications including survival modeling, immune phenotyping, biomarker discovery, and the development of new computational pathology methods.

\section{Related Work}
\label{sec:related_work}

The composition of the TME --- including the spatial organization of cancer cells, immune infiltrates, and stromal components --- has emerged as a key determinant of disease progression, prognosis, and treatment response~\cite{saltz2018,wang2020,raczkowska2022}. 
Routine H\&E-stained WSIs capture rich morphological information about the TME, and a growing body of work has demonstrated that AI-derived tissue and cell features from H\&E can predict survival outcomes, gene mutations, and molecular biomarkers across cancer types~\cite{diao2021,arslan2024,lee2022}. 
Notably, Saltz et~al.~\cite{saltz2018} generated computational TIL maps across 13 TCGA tumor types, establishing an early large-scale public resource linking H\&E-derived immune features to molecular and clinical phenotypes. 
However, that work focused on a single cell type (lymphocytes) at the patch level, without fine-grained cell classification or spatial neighborhood analysis. 

On the methodological side, automated cell detection and classification in H\&E has advanced substantially. 
HoVer-Net~\cite{graham2019} introduced simultaneous nuclear segmentation and classification, and PanNuke~\cite{gamper2019,gamper2020} provided an early large-scale pan-cancer benchmark for nuclei instance segmentation across 19 tissue types. 
More recently, CellViT~\cite{hoerst2024} combined vision transformers with foundation model backbones to achieve state-of-the-art performance on PanNuke, and HistoPLUS~\cite{adjadj2025} further improved classification of rare and clinically relevant cell types using a compact, pathology-specific encoder. 
While these methods provide powerful open-source tools for cell-level analysis, applying them at scale to generate ready-to-use, cohort-level TME profiles still requires substantial computational infrastructure, domain expertise, and careful curation of pathologist annotations. 

OpenTME addresses this gap by providing pre-computed, comprehensive TME characterization --- spanning tissue segmentation, eleven cell classes, tumor zones, TLSs, and spatial neighborhood features --- uniformly applied across 5,268 TCGA slides from ten cancer types.
Unlike patch-level or single-cell-type resources, OpenTME offers a multi-layered, quantitative description of the TME at single-cell resolution, directly usable for downstream analyses such as biomarker discovery or survival modeling without requiring users to run AI inference pipelines themselves.

\section{Dataset Description}
\label{sec:dataset}

This section describes the two components underlying OpenTME: the HETA application used to generate all outputs (Section~\ref{ssec:heta}), and the OpenTME dataset itself, including its scope, contents, and access (Section~\ref{ssec:opentme}).

\subsection{HETA}
\label{ssec:heta}

HETA is an AI-powered application developed by Aignostics for comprehensive, high-quality spatial profiling of the tumor microenvironment in H\&E-stained WSIs at single-cell resolution. 
It is built on the Atlas family of pathology foundation models \cite{dippel2024,alber2025,alber2026}, co-developed by Aignostics, Charité – Universitätsmedizin Berlin, LMU Munich, and Mayo Clinic. 
The underlying foundation model is a vision transformer (ViT) pre-trained on a large-scale, diverse histopathology dataset using self-supervised learning with the DINO framework \cite{oquab2023,simeoni2025}. 
Task-specific models were obtained by fine-tuning the Atlas backbone in a supervised fashion on pathologist-informed annotations across diverse training data.

HETA applies a sequential six-stage workflow to each WSI:

\paragraph{(1) Tissue Quality Control (QC)} segments the slide into regions of valid tissue, out-of-focus areas, tissue artifacts, and pen marker regions, ensuring that only tissue of sufficient quality is passed to downstream analysis. The QC model performs pixel-wise classification using a multi-scale semantic segmentation architecture on top of the ViT foundation model backbone.

\paragraph{(2) Tissue Segmentation} classifies valid tissue into nine distinct tissue types: carcinoma, (normal) epithelial tissue, stroma, necrosis, blood, vessel, nerve, fat, and other. The tissue segmentation model shares the same semantic segmentation architecture as the QC model.

\paragraph{(3) Cell Detection and Classification} identifies individual cells using a custom StarDist-based \cite{schmidt2018} nucleus segmentation model applied within segmented tissue regions, excluding necrotic areas. Each detected cell is then classified into one of eleven classes --- carcinoma cell, (normal) epithelial cell, fibroblast, lymphocyte, plasma cell, macrophage, neutrophil, eosinophil, endothelial cell, mitosis, and other --- by a multi-scale classifier on top of the foundation model.

\paragraph{(4) Tumor Zones} partition valid tissue by position relative to the tumor boundary into tumor core, inner and outer invasive margin (each extending 500~\textmu m from the boundary), and extratumoral tissue.

\paragraph{(5) Tertiary Lymphoid Structure (TLS) Detection} segments TLSs and classifies each as immature or mature, characterizing it by its shape, vessel content, and cell composition.

\paragraph{(6) Readouts} are computed based on all classified cells, tissue regions, tumor zones, and TLSs, further combined into neighborhood features. Readouts include cell counts, ratios, densities, and nuclear morphology features per cell type and are reported both at the slide level and stratified by tissue type and tumor zone. Neighborhood analysis further quantifies spatial relationships between cell types by computing co-occurrence statistics, ratios, and densities within defined radii of 20~\textmu m and 40~\textmu m. In addition, curated features readily capture key biological characteristics of the TME, such as tumor-infiltrating lymphocytes (TILs), tumor mitotic index, mature TLS counts, or tumor-stroma ratio.

\begin{figure}[tbh]
\centering
\setlength{\fboxsep}{0em}


\begin{minipage}[t]{0.245\linewidth}
    \centering \sffamily
    \fbox{\includegraphics[width=0.95\linewidth]{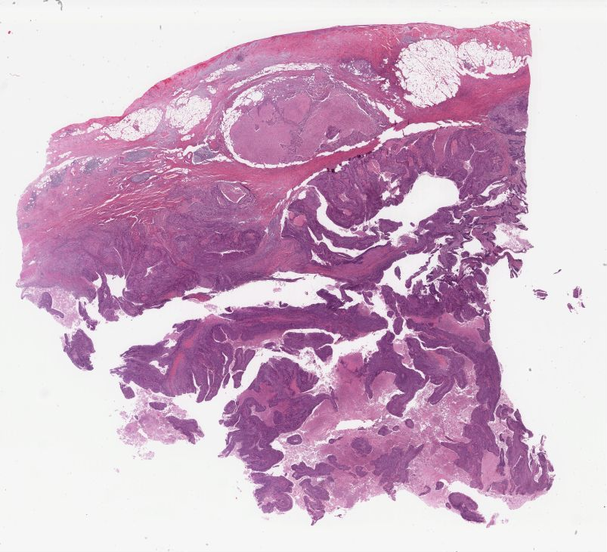}}\\[0.3em]
    \fbox{\includegraphics[width=0.95\linewidth]{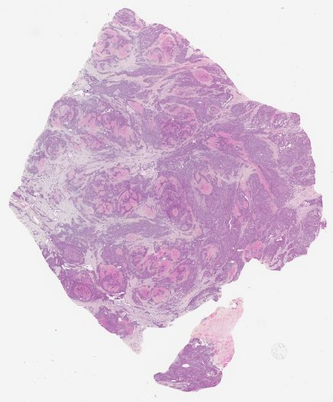}}\\[0.4em]
    WSI
\end{minipage}%
\begin{minipage}[t]{0.03\linewidth}
    \centering
    \hspace{0.99\linewidth}
\end{minipage}%
\begin{minipage}[t]{0.245\linewidth}
    \centering \sffamily
    \fbox{\includegraphics[width=0.95\linewidth]{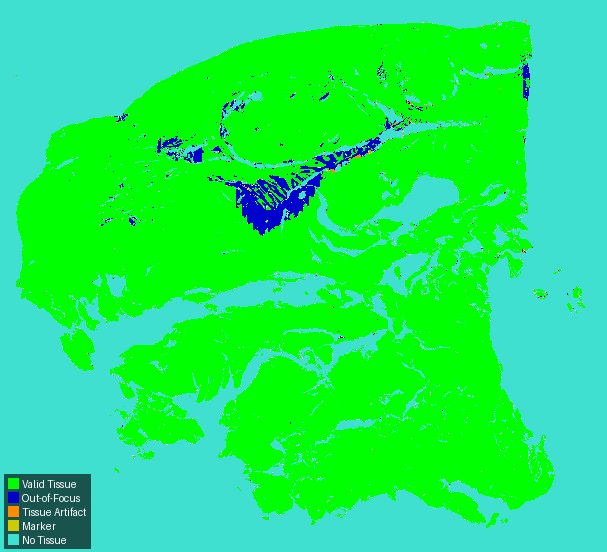}}\\[0.3em]
    \fbox{\includegraphics[width=0.95\linewidth]{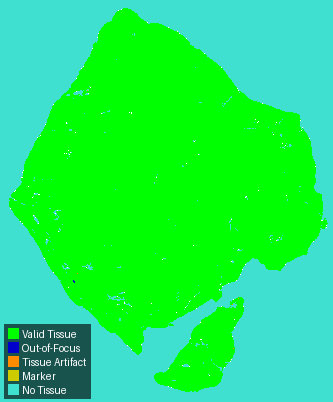}}\\[0.4em]
    Tissue QC
\end{minipage}%
\begin{minipage}[t]{0.245\linewidth}
    \centering \sffamily
    \fbox{\includegraphics[width=0.95\linewidth]{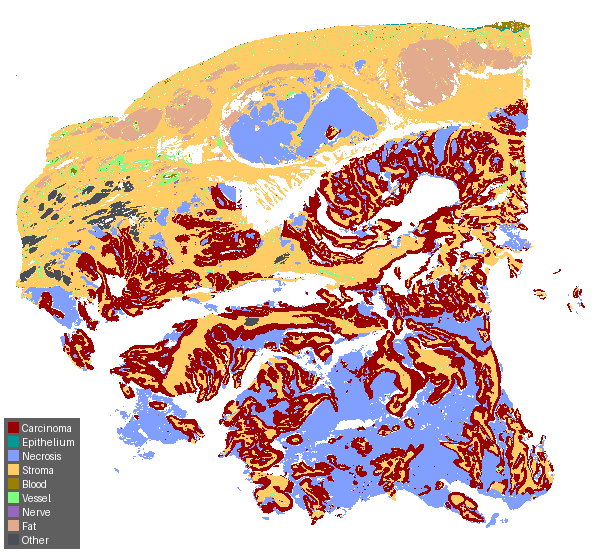}}\\[0.3em]
    \fbox{\includegraphics[width=0.95\linewidth]{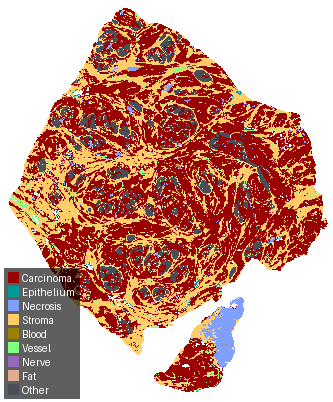}}\\[0.4em]
    Tissue Segmentation
\end{minipage}%
\begin{minipage}[t]{0.245\linewidth}
    \centering \sffamily
    \fbox{\includegraphics[width=0.95\linewidth]{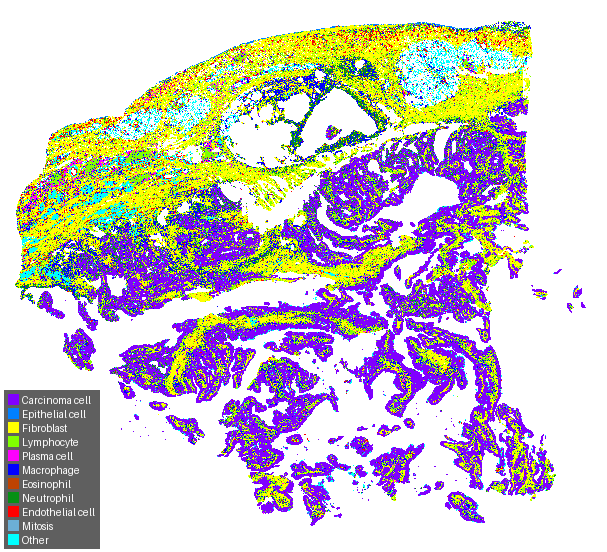}}\\[0.3em]
    \fbox{\includegraphics[width=0.95\linewidth]{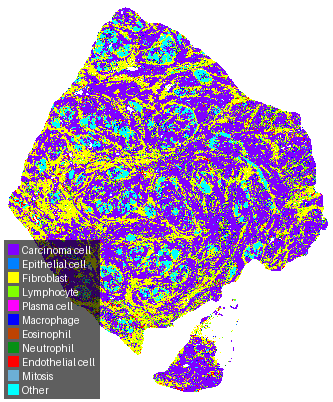}}\\[0.4em]
    Cell Classification
\end{minipage}
\caption{Two TCGA examples from the OpenTME dataset with HETA Tissue QC, Tissue Segmentation, and Cell Classification model outputs. Model outputs are integrated to derive over 20,000 quantitative readouts per slide at cell-level resolution, including spatial neighborhood features.}
\label{fig:example_thumbnails}
\end{figure}

Examples of HETA model outputs are shown in Figure~\ref{fig:example_thumbnails}. Detailed descriptions of the model classes are provided in the HETA validation study~\cite{standvoss2026}\footnote{HETA was previously referred to as Atlas H\&E-TME.} and in the OpenTME User Guide~\cite{opentme_userguide}.

HETA currently supports ten cancer types: bladder, breast, colorectal, head and neck, liver/biliary (hepatocellular carcinoma and cholangiocarcinoma), lung (NSCLC and lung neuroendocrine tumors), ovary, pancreas, prostate, and stomach. The application has been extensively validated against board-certified pathologist annotations across datasets from multiple sources and scanner types, covering at least 90\% of invasive morphological subtypes per supported cancer type, though it can be applied more broadly within these indications.
In the validation study~\cite{standvoss2026}, HETA cell classification matched or exceeded the performance of expert pathologists working from H\&E alone when benchmarked against an IHC-informed multi-pathologist consensus reference. An in-breadth evaluation on over 200,000 pathologist annotations from more than 1,500 cases across the covered cancer types and their common metastatic sites further demonstrated consistent performance across morphological and technical variation. 

\subsection{The OpenTME Dataset}
\label{ssec:opentme}

OpenTME provides pre-computed, quantitative tumor microenvironment profiles for H\&E-stained diagnostic WSIs from TCGA. 
By applying HETA uniformly across a large, publicly available cohort, OpenTME offers a consistent, AI-derived characterization of the TME that can serve as a foundation for downstream research including biomarker discovery, survival analysis, and spatial biology.

\paragraph{Data source and slide selection.}
All diagnostic formalin-fixed paraffin-embedded (FFPE) slides from TCGA projects corresponding to ten cancer types (bladder, breast, colorectal, head and neck, liver, lung, ovarian, pancreatic, prostate, and stomach cancer) were sourced via the NCI Genomic Data Commons (GDC).
From an initial set of 5,363 slides, 95 were excluded: 91 due to missing resolution metadata (mpp) or file corruption,
3 non-H\&E slides (2 IHC, 1 Masson's trichrome), and 1 slide that did not pass quality control due to full out-of-focus tissue.
The final OpenTME v2.0 dataset comprises 5,268 slides across ten indications and thirteen TCGA projects, as detailed in Table~\ref{tab:tcga_slides}.

\begin{table}[tbh]
  \centering
  \caption{TCGA slides included in OpenTME.}
  \label{tab:tcga_slides}
  \input{tab/tcga_slides}
\end{table}

\paragraph{Dataset contents.}
For each slide, OpenTME provides over 20,000 quantitative readouts in CSV format, organized at the slide level in coherently structured tables with one file per readout table and cancer type.
Curated features readily provide key biological characteristics of the TME in four categories: tumor composition (e.g., tumor-stroma ratio, tumor cellularity), cell morphology and proliferation (e.g., tumor mitotic index), immune contexture (e.g., TILs, mature TLS counts), and vascularization.
They are derived from the raw HETA readouts, which span five categories: (i) tissue quality control metrics, reporting area and relative coverage for each QC region; (ii) tissue segmentation and tumor zone metrics, including absolute and relative area, region count, and polygon morphology descriptors for each of the nine tissue types, the four tumor zones, and their combinations; (iii) cell metrics, including count, fraction, density, and nuclear morphology (area, circularity, eccentricity) per cell class, reported both at slide level and stratified by tissue type and tumor zone; (iv) neighborhood metrics, capturing spatial co-occurrence statistics, fractions, densities, and distances between cells within 20~\textmu m and 40~\textmu m radii, stratified by tissue type and tumor zone; and (v) TLS metrics, including counts, densities, morphology, and cell composition per TLS class.
A complete overview of all readouts is provided in Appendix~\ref{app:readouts}.

In addition to quantitative readouts, each slide is accompanied by thumbnail images with overlays for tissue QC, tissue segmentation, and cell classification predictions, enabling visual inspection and quality assessment (see Figure~\ref{fig:example_thumbnails}). 


\paragraph{Access and licensing.}
OpenTME is available as a gated dataset on Hugging Face.\footnote{\url{https://huggingface.co/datasets/Aignostics/OpenTME}} 
Access requires individual registration with an institutional email address and is granted at Aignostics' discretion. 
The dataset is released for non-commercial academic research use only; it may not be used for diagnostic purposes, model training aimed at replicating HETA capabilities, or redistribution. 
All users must additionally comply with TCGA data use policies. 
Full license terms are provided alongside the dataset. 

\paragraph{TME Studio.}
To facilitate exploration and analysis of OpenTME, we provide TME Studio, a collection of interactive marimo notebooks.\footnote{\url{https://github.com/aignostics/tme-studio}} 
TME Studio includes tutorials for loading and exploring the data, example analyses such as immune infiltrate classification and Kaplan–Meier survival analysis, and a demo notebook showcasing selected visualizations. 
TME Studio is intended as a starting point; users are encouraged to build on it using the broader ecosystem of open-source analysis tools.

\section{Conclusion}
\label{sec:conclusion}

We present OpenTME, an open-access dataset of quantitative tumor microenvironment profiles derived from 5,268 TCGA WSIs across ten cancer types, generated using HETA.
By providing consistent, cell-level TME characterization in a ready-to-use format, OpenTME removes a key barrier to large-scale H\&E-based spatial biology research and enables downstream applications ranging from biomarker discovery to survival modeling.
Alongside the dataset, the HETA Research Access Program offers academic researchers the ability to apply for access to the same analysis pipeline for their own data.

The current release has some limitations.
OpenTME covers ten cancer types, which, while representing common solid tumors, do not yet span the full breadth of malignancies available in TCGA.
The readouts provided are aggregated at the slide level; spatially resolved outputs such as cell coordinates and polygon geometries are not included in the public dataset but can be requested through the Research Access Program. 
Furthermore, as with any AI-derived dataset, the outputs reflect the current performance characteristics of the underlying models and 
should be interpreted accordingly. 

We view OpenTME as a living resource. 
Future releases will expand cancer type coverage, introduce more granular and extensive readouts, and incorporate 
improvements from updated versions of the Atlas foundation model family and the HETA application.
We invite the research community to build on OpenTME and welcome input and feedback to its continued development.

\section*{Acknowledgments}
\label{sec:acks}
We would like to thank the teams at Aignostics, the Department of Pathology at 
Charité -- Universitätsmedizin Berlin, and the Department of Pathology 
at LMU Munich for their contributions to the development and validation 
of HETA.

The results published here are in whole or part based upon data generated by the TCGA Research Network: \url{https://www.cancer.gov/tcga}.

\bibliographystyle{plain} 
\bibliography{references} 

\newpage
\appendix



\section{HETA: Readouts}
\label{app:readouts}

OpenTME readouts are organized in long-format tables: each row corresponds to one combination of key columns (e.g., \texttt{CELL\_TYPE} and \texttt{TISSUE\_TYPE}), and the value columns hold the respective measurements.
In the readouts provided on Hugging Face, the rows of all slides of one indication are combined into one file per table, and each row is prefixed with the TCGA identifiers \texttt{TCGA\_FILE\_NAME}, \texttt{TCGA\_SLIDE\_UUID}, \texttt{TCGA\_CASE\_ID}, and \texttt{TCGA\_PROJECT\_ID}, as well as the \texttt{INDICATION}. 
Areas of regions are reported in mm\textsuperscript{2}, nucleus areas in \textmu m\textsuperscript{2}, lengths and coordinates in \textmu m, densities per mm\textsuperscript{2}, and fractions as values between 0 and 1. 
The whole tumor region comprises the tumor core and the inner invasive margin. 
Table~\ref{tab:readout_files} gives an overview of all readout tables, and Table~\ref{tab:curated_features} lists all curated features. 
Complete column definitions are provided in the OpenTME User Guide~\cite{opentme_userguide}.

\input{tab/readouts_files}

\input{tab/readouts_curated}

\end{document}

%% file: tab/tcga_slides.tex
\begin{tabular}{@{}llrrr@{}}
    \toprule
    Cancer type & TCGA project(s)   & Cases & Slides    & Slides (final) \\
    \midrule
    Bladder     & TCGA-BLCA         & 386   & 457       & 457  \\
    Breast      & TCGA-BRCA         & 1062  & 1133      & 1125 \\
    \multirow{2}{*}{Colorectal}
                & TCGA-COAD         & 451   & 459       & 442  \\
                & TCGA-READ         & 165   & 166       & 158  \\
    Head and neck & TCGA-HNSC       & 450   & 472       & 472  \\
    \multirow{2}{*}{Liver}
                & TCGA-LIHC         & 365   & 379       & 372  \\
                & TCGA-CHOL         & 39    & 39        & 39   \\
    \multirow{2}{*}{Lung}
                & TCGA-LUAD         & 478   & 541       & 529  \\
                & TCGA-LUSC         & 478   & 512       & 512  \\
    Ovary       & TCGA-OV           & 105   & 105       & 105  \\
    Pancreas    & TCGA-PAAD         & 183   & 209       & 209  \\
    Prostate    & TCGA-PRAD         & 403   & 449       & 448  \\
    Stomach     & TCGA-STAD         & 416   & 442       & 400  \\
    \midrule
    \textbf{Total} &                & \textbf{4981} & \textbf{5363} & \textbf{5268} \\
\bottomrule
\end{tabular}

%% file: tab/readouts_files.tex
{\small
\begin{longtable}{@{}>{\raggedright\arraybackslash}p{4.4cm}>{\raggedright\arraybackslash}p{2.5cm}>{\raggedright\arraybackslash}p{6.2cm}@{}}
\caption{Overview of readout tables. Each table is provided as a separate CSV file (\texttt{<table>.csv}).}
\label{tab:readout_files} \\
\toprule
Table & One row per & Value columns \\
\midrule
\endfirsthead
\caption[]{Overview of readout tables (\emph{continued}).} \\
\toprule
Table & One row per & Value columns \\
\midrule
\endhead
\midrule
\multicolumn{3}{r}{\emph{Continued on next page}} \\
\bottomrule
\endfoot
\bottomrule
\multicolumn{3}{@{}p{13.1cm}@{}}{\footnotesize
$^{a}$~\ttb{NUCLEUS\_MEAN\_AREA}, \ttb{NUCLEUS\_STD\_AREA}, \ttb{NUCLEUS\_MEAN\_CIRCULARITY}, \ttb{NUCLEUS\_STD\_CIRCULARITY}, \ttb{NUCLEUS\_MEAN\_ECCENTRICITY}, \ttb{NUCLEUS\_STD\_ECCENTRICITY}.
$^{b}$~Region in which the center cells are located: each tissue type except necrosis, and each tumor zone.
$^{c}$~Immature TLS, mature TLS, and all TLSs.
$^{d}$~Means over all TLSs of the per-TLS measurements listed for \ttb{tertiary\_lymphoid\_structure\_stats}.} \\
\endlastfoot

\multicolumn{3}{@{}l}{\emph{Curated features (see Table~\ref{tab:curated_features})}} \\
\midrule
\ttb{curated\_tumor\_composition\_features}\newline
\ttb{curated\_cell\_morphology\_proliferation\_features}\newline
\ttb{curated\_immune\_contexture\_features}\newline
\ttb{curated\_vascularization\_features} &
  feature &
  \ttb{FEATURE}, \ttb{VALUE}, \ttb{UNIT}, \ttb{DESCRIPTION}, \ttb{DATA\_SOURCE} \\

\midrule
\multicolumn{3}{@{}l}{\emph{Raw HETA results}} \\
\midrule
\ttb{whole\_slide\_stats} &
  slide &
  \ttb{IMAGE\_RESOLUTION}, \ttb{AREA\_ABSOLUTE}, \ttb{CELL\_COUNT} \\
\ttb{qc\_stats} &
  QC region &
  \ttb{AREA\_ABSOLUTE}, \ttb{AREA\_FRACTION} \\
\ttb{cell\_aggregates} &
  cell class &
  \ttb{CELL\_COUNT}, \ttb{CELL\_FRACTION}, \ttb{CELL\_DENSITY}, nucleus morphometry$^{a}$ \\
\ttb{tissue\_segmentation\_stats} &
  tissue type &
  \ttb{AREA\_ABSOLUTE}, \ttb{AREA\_FRACTION}, \ttb{REGION\_COUNT}, \ttb{MEAN\_MINOR\_AXIS}, \ttb{MEAN\_PERIMETER} \\
\ttb{tissue\_segmentation\_cell\_aggregates} &
  tissue type $\times$ cell class &
  \ttb{CELL\_COUNT}, \ttb{CELL\_FRACTION}, \ttb{CELL\_DENSITY} \\
\ttb{tumor\_zone\_stats} &
  tumor zone &
  \ttb{AREA\_ABSOLUTE}, \ttb{AREA\_FRACTION}, \ttb{REGION\_COUNT}, \ttb{MEAN\_MINOR\_AXIS}, \ttb{MEAN\_PERIMETER} \\
\ttb{tumor\_zone\_cell\_aggregates} &
  tumor zone $\times$ cell class &
  \ttb{CELL\_COUNT}, \ttb{CELL\_FRACTION}, \ttb{CELL\_DENSITY}, nucleus morphometry$^{a}$ \\
\ttb{tissue\_segmentation\_and\_tumor\_zone\_stats} &
  tissue type $\times$ tumor zone &
  \ttb{AREA\_ABSOLUTE}, \ttb{AREA\_FRACTION}, \ttb{REGION\_COUNT}, \ttb{MEAN\_MINOR\_AXIS}, \ttb{MEAN\_PERIMETER} \\
\ttb{tissue\_segmentation\_and\_tumor\_zone\_cell\_aggregates} &
  tissue type $\times$ tumor zone $\times$ cell class &
  \ttb{CELL\_COUNT}, \ttb{CELL\_FRACTION}, \ttb{CELL\_DENSITY}, nucleus morphometry$^{a}$ \\
\ttb{cell\_neighborhood\_stats} &
  center cell class $\times$ target cell class $\times$ region$^{b}$ $\times$ radius (20, 40~\textmu m) &
  \ttb{TOTAL\_NEIGHBOR\_COUNT}, \ttb{CENTER\_CELLS\_WITH\_TARGET\_NEIGHBOR}, \ttb{TARGET\_NEIGHBOR\_COUNT}, \ttb{TARGET\_NEIGHBOR\_DENSITY}, \ttb{TARGET\_NEIGHBOR\_FRACTION}, \ttb{MEAN\_MIN\_DISTANCE\_TO\_TARGET\_NEIGHBOR} \\
\ttb{tertiary\_lymphoid\_structure\_slide\_stats} &
  TLS class$^{c}$ &
  \ttb{TLS\_COUNT}, \ttb{TLS\_DENSITY}, \ttb{AREA\_ABSOLUTE}, mean TLS morphology and vessel content$^{d}$ \\
\ttb{tertiary\_lymphoid\_structure\_slide\_cell\_aggregates} &
  TLS class$^{c}$ $\times$ cell class &
  \ttb{MEAN\_CELL\_COUNT}, \ttb{MEAN\_CELL\_FRACTION}, \ttb{MEAN\_CELL\_DENSITY} \\

\midrule
\multicolumn{3}{@{}l}{\emph{Granular readouts (available through the HETA Research Access Program)}} \\
\midrule
\ttb{cell\_stats} &
  cell &
  \ttb{CELL\_ID}, \ttb{CELL\_CENTER\_X}, \ttb{CELL\_CENTER\_Y}, \ttb{CELL\_TYPE}, \ttb{IN\_TISSUE\_TYPE}, \ttb{IN\_TUMOR\_ZONE}, \ttb{NUCLEUS\_AREA}, \ttb{NUCLEUS\_CIRCULARITY}, \ttb{NUCLEUS\_ECCENTRICITY} \\
\ttb{tertiary\_lymphoid\_structure\_stats} &
  TLS &
  \ttb{TLS\_ID}, \ttb{TLS\_CLASS}, \ttb{TLS\_CENTER\_X}, \ttb{TLS\_CENTER\_Y}, \ttb{AREA\_ABSOLUTE}, \ttb{CONVEX\_AREA}, \ttb{CIRCULARITY}, \ttb{ECCENTRICITY}, \ttb{SOLIDITY}, \ttb{MAJOR\_AXIS}, \ttb{MINOR\_AXIS}, \ttb{AREA\_VESSEL}, \ttb{REGION\_COUNT\_VESSEL} \\
\ttb{tertiary\_lymphoid\_structure\_cell\_aggregates} &
  TLS $\times$ cell class &
  \ttb{CELL\_COUNT}, \ttb{CELL\_FRACTION}, \ttb{CELL\_DENSITY} \\

\end{longtable}
}

%% file: tab/readouts_curated.tex
{\small
\begin{longtable}{@{}>{\raggedright\arraybackslash}p{5.6cm}>{\raggedright\arraybackslash}p{1.6cm}>{\raggedright\arraybackslash}p{5.9cm}@{}}
\caption{Curated features. Unless stated otherwise, features refer to the whole tumor region (tumor core and inner invasive margin).}
\label{tab:curated_features} \\
\toprule
Feature & Unit & Description \\
\midrule
\endfirsthead
\caption[]{Curated features (\emph{continued}).} \\
\toprule
Feature & Unit & Description \\
\midrule
\endhead
\midrule
\multicolumn{3}{r}{\emph{Continued on next page}} \\
\bottomrule
\endfoot
\bottomrule
\endlastfoot

\multicolumn{3}{@{}l}{\emph{Tumor composition (\ttb{curated\_tumor\_composition\_features})}} \\
\midrule
\ttb{TUMOR\_STROMA\_RATIO} & ratio &
  Stroma area relative to the sum of carcinoma and stroma area. \\
\ttb{TUMOR\_STROMA\_RATIO\_INNER\_INVASIVE\_MARGIN} & ratio &
  Stroma area relative to the sum of carcinoma and stroma area in the inner invasive margin. \\
\ttb{NECROSIS\_COVERAGE} & fraction &
  Necrotic area as a fraction of the whole tumor region. \\
\ttb{CARCINOMA\_CELL\_COUNT} & count &
  Number of carcinoma cells on the slide. \\
\ttb{CARCINOMA\_NUCLEAR\_AREA} & mm\textsuperscript{2} &
  Summed nuclear area of all carcinoma cells on the slide. \\
\ttb{TUMOR\_CELLULARITY\_TOTAL\_TISSUE} & fraction &
  Carcinoma cells as a fraction of all cells in the tissue. \\
\ttb{TUMOR\_CELLULARITY\_WHOLE\_TUMOR\_REGION} & fraction &
  Carcinoma cells as a fraction of all cells in the whole tumor region. \\
\ttb{CARCINOMA\_NUCLEAR\_AREA\_FRACTION\_TOTAL\_TISSUE} & fraction &
  Carcinoma nuclear area as a fraction of the total nuclear area in the tissue. \\
\ttb{CARCINOMA\_NUCLEAR\_AREA\_FRACTION\_WHOLE\_TUMOR\_REGION} & fraction &
  Carcinoma nuclear area as a fraction of the total nuclear area in the whole tumor region. \\
\ttb{FIBROBLAST\_DENSITY} & cells/mm\textsuperscript{2} &
  Fibroblasts per mm\textsuperscript{2} of stroma. \\
\ttb{FIBROBLAST\_FRACTION} & fraction &
  Fibroblasts as a fraction of all cells in stroma. \\

\midrule
\multicolumn{3}{@{}l}{\emph{Cell morphology and proliferation (\ttb{curated\_cell\_morphology\_proliferation\_features})}} \\
\midrule
\ttb{CARCINOMA\_NUCLEUS\_AREA\_MEAN} & \textmu m\textsuperscript{2} &
  Mean nuclear area of carcinoma cells on the slide. \\
\ttb{CARCINOMA\_NUCLEUS\_AREA\_CV} & ratio &
  Coefficient of variation of the nuclear area of carcinoma cells on the slide. \\
\ttb{FIBROBLAST\_NUCLEUS\_AREA\_MEAN} & \textmu m\textsuperscript{2} &
  Mean nuclear area of fibroblasts in stroma. \\
\ttb{MITOTIC\_COUNT} & cells/mm\textsuperscript{2} &
  Number of mitoses per mm\textsuperscript{2} of carcinoma. \\
\ttb{MITOTIC\_INDEX} & fraction &
  Number of mitoses as a fraction of carcinoma cells. \\

\midrule
\multicolumn{3}{@{}l}{\emph{Immune contexture (\ttb{curated\_immune\_contexture\_features})}} \\
\midrule
\ttb{MATURE\_TLS\_COUNT} & count &
  Number of mature TLSs on the slide. \\
\ttb{MATURE\_TLS\_DENSITY} & per mm\textsuperscript{2} &
  Mature TLSs per mm\textsuperscript{2} of tissue (excluding necrosis and blood). \\
\ttb{IMMATURE\_TLS\_COUNT} & count &
  Number of immature TLSs on the slide. \\
\ttb{IMMATURE\_TLS\_DENSITY} & per mm\textsuperscript{2} &
  Immature TLSs per mm\textsuperscript{2} of tissue (excluding necrosis and blood). \\
\ttb{TIL\_FRACTION\_CARCINOMA} & fraction &
  Lymphocytes as a fraction of all cells in carcinoma (intratumoral TILs). \\
\ttb{TIL\_FRACTION\_STROMA} & fraction &
  Lymphocytes as a fraction of all cells in stroma (stromal TILs). \\
\ttb{TIL\_DENSITY\_CARCINOMA} & cells/mm\textsuperscript{2} &
  Lymphocytes per mm\textsuperscript{2} of carcinoma (intratumoral TILs). \\
\ttb{TIL\_DENSITY\_STROMA} & cells/mm\textsuperscript{2} &
  Lymphocytes per mm\textsuperscript{2} of stroma (stromal TILs). \\
\ttb{TIL\_COMPARTMENT\_EXCLUSION\_INDEX} & ratio &
  Stromal over intratumoral TIL density. \\
\ttb{TIL\_DENSITY\_TUMOR\_CORE} & cells/mm\textsuperscript{2} &
  Lymphocytes per mm\textsuperscript{2} in the tumor core. \\
\ttb{TIL\_DENSITY\_INVASIVE\_MARGIN} & cells/mm\textsuperscript{2} &
  Lymphocytes per mm\textsuperscript{2} in the inner and outer invasive margin. \\
\ttb{TIL\_DENSITY\_OUTER\_INVASIVE\_MARGIN} & cells/mm\textsuperscript{2} &
  Lymphocytes per mm\textsuperscript{2} in the outer invasive margin. \\
\ttb{REFERENCE\_PERCENTILE\_TIL\_DENSITY\_STROMA} & percent &
  Percentile of the stromal TIL density relative to the pan-cancer TCGA reference distribution. \\
\ttb{REFERENCE\_PERCENTILE\_TIL\_DENSITY\_CARCINOMA} & percent &
  Percentile of the intratumoral TIL density relative to the pan-cancer TCGA reference distribution. \\
\ttb{REFERENCE\_PERCENTILE\_TIL\_DENSITY\_TUMOR\_CORE} & percent &
  Percentile of the tumor core TIL density relative to the pan-cancer TCGA reference distribution. \\
\ttb{REFERENCE\_PERCENTILE\_TIL\_DENSITY\_INVASIVE\_MARGIN} & percent &
  Percentile of the invasive margin TIL density relative to the pan-cancer TCGA reference distribution. \\
\ttb{REFERENCE\_PERCENTILE\_TIL\_DENSITY\_OUTER\_INVASIVE\_MARGIN} & percent &
  Percentile of the outer invasive margin TIL density relative to the pan-cancer TCGA reference distribution. \\
\ttb{MEAN\_REFERENCE\_PERCENTILE\_TIL\_DENSITY\_MARGIN\_CORE} & percent &
  Mean of the tumor core and invasive margin reference percentiles; an H\&E-based approximation of the CD3/CD8 Immunoscore. \\
\ttb{TIL\_MARGIN\_EXCLUSION\_INDEX} & ratio &
  Invasive margin over tumor core TIL density. \\

\midrule
\multicolumn{3}{@{}l}{\emph{Vascularization (\ttb{curated\_vascularization\_features})}} \\
\midrule
\ttb{VESSEL\_AREA\_FRACTION\_IN\_WHOLE\_TUMOR\_REGION} & fraction &
  Vessel area as a fraction of the whole tumor region (excluding necrosis and blood). \\
\ttb{MEAN\_VESSEL\_PERIMETER\_IN\_WHOLE\_TUMOR\_REGION} & \textmu m &
  Mean perimeter of vessels. \\
\ttb{MEAN\_VESSEL\_MINOR\_AXIS\_IN\_WHOLE\_TUMOR\_REGION} & \textmu m &
  Mean minor axis of vessels (vessel caliber). \\
\ttb{ENDOTHELIAL\_CELL\_DENSITY\_IN\_WHOLE\_TUMOR\_REGION} & cells/mm\textsuperscript{2} &
  Endothelial cells per mm\textsuperscript{2} of the whole tumor region (excluding necrosis and blood). \\
\ttb{ENDOTHELIAL\_CELL\_DENSITY\_IN\_CARCINOMA} & cells/mm\textsuperscript{2} &
  Endothelial cells per mm\textsuperscript{2} of carcinoma. \\

\end{longtable}
}

%% file: references.bib
@article{dippel2024,
    title = {{RudolfV}: {A} {Foundation} {Model} by {Pathologists} for {Pathologists}},
    author = {Dippel, Jonas and Feulner, Barbara and Winterhoff, Tobias and Milbich, Timo and Tietz, Stephan and Schallenberg, Simon and Dernbach, Gabriel and Kunft, Andreas and Heinke, Simon and Eich, Marie-Lisa and Ribbat-Idel, Julika and Krupar, Rosemarie and Anders, Philipp and Prenißl, Niklas and Jurmeister, Philipp and Horst, David and Ruff, Lukas and Müller, Klaus-Robert and Klauschen, Frederick and Alber, Maximilian},
    journal={arXiv preprint arXiv:2401.04079},
    year = {2024}
}

@article{alber2025,
    title={Atlas: A Novel Pathology Foundation Model by Mayo Clinic, Charit\'e, and Aignostics}, 
    author={Maximilian Alber and Stephan Tietz and Jonas Dippel and Timo Milbich and Timothée Lesort and Panos Korfiatis and Moritz Krügener and Beatriz Perez Cancer and Neelay Shah and Alexander Möllers and Philipp Seegerer and Alexandra Carpen-Amarie and Kai Standvoss and Gabriel Dernbach and Edwin de Jong and Simon Schallenberg and Andreas Kunft and Helmut Hoffer von Ankershoffen and Gavin Schaeferle and Patrick Duffy and Matt Redlon and Philipp Jurmeister and David Horst and Lukas Ruff and Klaus-Robert Müller and Frederick Klauschen and Andrew Norgan},
    journal={arXiv preprint arXiv:2501.05409},
    year={2025}
}

@article{alber2026,
    title={Atlas 2 -- Foundation models for clinical deployment}, 
    author={Maximilian Alber and Timo Milbich and Alexandra Carpen-Amarie and Stephan Tietz and Jonas Dippel and Lukas Muttenthaler and Beatriz Perez Cancer and Alessandro Benetti and Panos Korfiatis and Elias Eulig and Jérôme Lüscher and Jiasen Wu and Sayed Abid Hashimi and Gabriel Dernbach and Simon Schallenberg and Neelay Shah and Moritz Krügener and Aniruddh Jammoria and Jake Matras and Patrick Duffy and Matt Redlon and Philipp Jurmeister and David Horst and Lukas Ruff and Klaus-Robert Müller and Frederick Klauschen and Andrew Norgan},
    journal={arXiv preprint arXiv:2601.05148},
    year={2026}
}

@article{zimmermann2024,
    title={{Virchow2}: Scaling Self-Supervised Mixed Magnification Models in Pathology},
    author={Zimmermann, Eric and Vorontsov, Eugene and Viret, Julian and Casson, Adam and Zelechowski, Michal and Shaikovski, George and Tenenholtz, Neil and Hall, James and Klimstra, David and Yousfi, Razik and Fuchs, Thomas and Fusi, Nicol\`{o} and Liu, Siqi and Severson, Kristen},
    journal={arXiv preprint arXiv:2408.00738},
    year={2024},
}

@article{chen2024,
  title={Towards a General-Purpose Foundation Model for Computational Pathology},
  author={Chen, Richard J. and Ding, Tong and Lu, Ming Y. and Williamson, Drew F. K. and Jaume, Guillaume and Song, Andrew H. and Chen, Bowen and Zhang, Andrew and Shao, Daniel and Shaban, Muhammad and Williams, Mane and Oldenburg, Lukas and Weishaupt, Luca L. and Wang, Judy J. and Vaidya, Anurag and Le, Long Phi and Gerber, Georg and Sahai, Sharifa and Williams, Walt and Mahmood, Faisal},
  journal={Nature Medicine},
  volume={30},
  number = {3},
  pages={850--862},
  year={2024},
  doi={10.1038/s41591-024-02857-3},
}

@article{oquab2023,
    title={{DINO}v2: Learning Robust Visual Features without Supervision},
    author={Maxime Oquab and Timoth{\'e}e Darcet and Th{\'e}o Moutakanni and Huy V. Vo and Marc Szafraniec and Vasil Khalidov and Pierre Fernandez and Daniel HAZIZA and Francisco Massa and Alaaeldin El-Nouby and Mido Assran and Nicolas Ballas and Wojciech Galuba and Russell Howes and Po-Yao Huang and Shang-Wen Li and Ishan Misra and Michael Rabbat and Vasu Sharma and Gabriel Synnaeve and Hu Xu and Herve Jegou and Julien Mairal and Patrick Labatut and Armand Joulin and Piotr Bojanowski},
    journal={Transactions on Machine Learning Research},
    issn={2835-8856},
    year={2024}
}

@article{simeoni2025,
    title={{DINOv3}},
    author={Sim{\'e}oni, Oriane and Vo, Huy V. and Seitzer, Maximilian and Baldassarre, Federico and Oquab, Maxime and Jose, Cijo and Khalidov, Vasil and Szafraniec, Marc and Yi, Seungeun and Ramamonjisoa, Micha{\"e}l and Massa, Francisco and Haziza, Daniel and Wehrstedt, Luca and Wang, Jianyuan and Darcet, Timoth{\'e}e and Moutakanni, Th{\'e}o and Sentana, Leonel and Roberts, Claire and Vedaldi, Andrea and Tolan, Jamie and Brandt, John and Couprie, Camille and Mairal, Julien and J{\'e}gou, Herv{\'e} and Labatut, Patrick and Bojanowski, Piotr},
    journal={arXiv preprint arXiv:2508.10104},
    year={2025}
}

@inproceedings{schmidt2018,
    title={Cell detection with star-convex polygons},
    author={Schmidt, Uwe and Weigert, Martin and Broaddus, Coleman and Myers, Gene},
    booktitle={International conference on medical image computing and computer-assisted intervention},
    pages={265--273},
    year={2018},
    organization={Springer}
}

@article{saltz2018,
    title={Spatial Organization and Molecular Correlation of Tumor-Infiltrating Lymphocytes Using Deep Learning on Pathology Images},
    author={Saltz, Joel and Gupta, Rajarsi and Hou, Le and Kurc, Tahsin and Singh, Pankaj and Nguyen, Vu and Samaras, Dimitris and Shroyer, Kenneth R. and Zhao, Tianhao and Batiste, Rebecca and Van Arnam, John and {The Cancer Genome Atlas Research Network} and Shmulevich, Ilya and Rao, Arvind U. K. and Lazar, Alexander J. and Sharma, Ashish and Thorsson, V{\'e}steinn},
    journal={Cell Reports},
    volume={23},
    number={1},
    pages={181--193.e7},
    year={2018}
}

@article{wang2020,
    title={Computational Staining of Pathology Images to Study the Tumor Microenvironment in Lung Cancer},
    author={Wang, Shidan and Rong, Ruichen and Yang, Donghan M. and Fujimoto, Junya and Yan, Shirley and Cai, Ling and Yang, Lin and Luo, Danni and Behrens, Carmen and Parra, Edwin R. and Yao, Bo and Xu, Lin and Wang, Tao and Zhan, Xiaowei and Wistuba, Ignacio I. and Minna, John and Xie, Yang and Xiao, Guanghua},
    journal={Cancer Research},
    volume={80},
    number={10},
    pages={2056--2066},
    year={2020}
}

@article{raczkowska2022,
    title={Deep Learning-Based Tumor Microenvironment Segmentation is Predictive of Tumor Mutations and Patient Survival in Non-Small-Cell Lung Cancer},
    author={R{\k{a}}czkowska, Alicja and Pa{\'s}nik, Iwona and Kukie{\l}ka, Micha{\l} and Nico{\'s}, Marcin and Budzinska, Magdalena A. and Kucharczyk, Tomasz and Szumi{\l}o, Justyna and Krawczyk, Pawe{\l} and Crosetto, Nicola and Szczurek, Ewa},
    journal={BMC Cancer},
    volume={22},
    number={1},
    pages={1001},
    year={2022}
}

@article{arslan2024,
    title={A Systematic Pan-Cancer Study on Deep Learning-Based Prediction of Multi-Omic Biomarkers from Routine Pathology Images},
    author={Arslan, Salim and Schmidt, Julian and Bass, Cher and Mehrotra, Debapriya and Geraldes, Andre and Singhal, Shikha and Hense, Julius and Li, Xiusi and Raharja-Liu, Pandu and Maiques, Oscar and Kather, Jakob Nikolas and Pandya, Pahini},
    journal={Communications Medicine},
    volume={4},
    pages={48},
    year={2024}
}

@article{lee2022,
    title={Derivation of Prognostic Contextual Histopathological Features from Whole-Slide Images of Tumours via Graph Deep Learning},
    author={Lee, Yongju and Park, Jeong Hwan and Oh, Sohee and Shin, Kyoungseob and Sun, Jiyu and Jung, Minsun and Lee, Cheol and Kim, Hyojin and Chung, Jin-Haeng and Moon, Kyung Chul and Kwon, Sunghoon},
    journal={Nature Biomedical Engineering},
    volume={6},
    number={12},
    pages={1452--1466},
    year={2022}
}

@article{graham2019,
    title={Hover-Net: Simultaneous Segmentation and Classification of Nuclei in Multi-Tissue Histology Images},
    author={Graham, Simon and Vu, Quoc Dang and Raza, Shan E Ahmed and Azam, Ayesha and Tsang, Yee Wah and Kwak, Jin Tae and Rajpoot, Nasir},
    journal={Medical Image Analysis},
    volume={58},
    pages={101563},
    year={2019}
}

@inproceedings{gamper2019,
    title={{PanNuke}: An Open Pan-Cancer Histology Dataset for Nuclei Instance Segmentation and Classification},
    author={Gamper, Jevgenij and Alemi Koohbanani, Navid and Benet, Ksenija and Khuram, Ali and Rajpoot, Nasir},
    booktitle={European Congress on Digital Pathology},
    series={Lecture Notes in Computer Science},
    volume={11435},
    pages={11--19},
    year={2019},
    publisher={Springer}
}

@article{gamper2020,
    title={{PanNuke} Dataset Extension, Insights and Baselines},
    author={Gamper, Jevgenij and Alemi Koohbanani, Navid and Graham, Simon and Jahanifar, Mostafa and Khurram, Syed Ali and Azam, Ayesha and Hewitt, Katherine and Rajpoot, Nasir},
    journal={arXiv preprint arXiv:2003.10778},
    year={2020},
}

@article{hoerst2024,
    title={{CellViT}: Vision Transformers for Precise Cell Segmentation and Classification},
    author={H{\"o}rst, Fabian and Rempe, Moritz and Heine, Lukas and Seibold, Constantin and Keyl, Julius and Baldini, Giulia and Ugurel, Selma and Siveke, Jens and Gr{\"u}nwald, Barbara and Egger, Jan and Kleesiek, Jens},
    journal={Medical Image Analysis},
    volume={94},
    pages={103143},
    year={2024}
}

@article{adjadj2025,
    title={Towards Comprehensive Cellular Characterisation of {H\&E} Slides},
    author={Adjadj, Benjamin and Bannier, Pierre-Antoine and Horent, Guillaume and Mandela, Sebastien and Lyon, Aurore and Schutte, Kathryn and Marteau, Ulysse and Gaury, Valentin and Dumont, Laura and Mathieu, Thomas and Belbahri, Reda and Schmauch, Beno{\^i}t and Durand, Eric and Von Loga, Katharina and Gillet, Lucie},
    journal={arXiv preprint arXiv:2508.09926},
    year={2025}
}

@article{diao2021,
    title={Human-interpretable image features derived from densely mapped cancer pathology slides predict diverse molecular phenotypes},
    author={Diao, James A and Wang, Jason K and Chui, Wan Fung and Mountain, Victoria and Gullapally, Sai Chowdary and Srinivasan, Ramprakash and Mitchell, Richard N and Glass, Benjamin and Hoffman, Sara and Rao, Sudha K and others},
    journal={Nature communications},
    volume={12},
    number={1},
    pages={1613},
    year={2021},
    publisher={Nature Publishing Group UK London}
}

@article{standvoss2026,
    title={Atlas {H\&E-TME}: Scalable {AI}-Based Tissue Profiling at Expert Pathologist-Level Accuracy},
    author={Standvoss, Kai and H{\"a}gele, Miriam and Krupar, Rosemarie and Ribbat-Idel, Julika and Altsch{\"u}ler, Jennifer and Erdmann, Gerrit and Pinckaers, Hans and Ramberger, Evelyn and Drinkwitz, Madleen and N{\'a}rai, {\'A}d{\'a}m and M{\"o}llers, Alexander and Lingelbach, Katja and Kons, Sebastian and H{\"o}nig, Lukas and Adig{\"u}zel, Recepcan and Bai{\~a}o, Joana and Megina Gonzalo, Alberto and Teodorescu, Marius and Eich, Marie-Lisa and Chetta, Paolo and Merchant, Shakil and Aumiller, Verena and Schallenberg, Simon and Norgan, Andrew and M{\"u}ller, Klaus-Robert and Ruff, Lukas and Alber, Maximilian and Klauschen, Frederick},
    journal={arXiv preprint arXiv:2606.12346},
    year={2026}
}

@misc{opentme_userguide,
    title={{OpenTME} User Guide},
    author={{Aignostics}},
    howpublished={Provided alongside the dataset at \url{https://huggingface.co/datasets/Aignostics/OpenTME}},
    year={2026}
}
